\documentclass[sigconf,nonacm,balance=false]{acmart}

\usepackage{tikz}
\usepackage{algorithm}
\usepackage{algorithmic}
\usepackage{array}
\usepackage{multirow}
\usepackage{caption}
\usepackage{adjustbox}
\usepackage{booktabs}
\usepackage{cellspace}
\usepackage{tabularx}
\usepackage{hhline}
\usepackage{makecell}
\usepackage{subcaption}
\usepackage{mdwmath}
\usepackage{mdwtab}
\usepackage{eqparbox}
\usepackage{stfloats}
\usepackage{amsmath}
\usepackage{bm}

\begin{document}

\title{Phantom Navigator: Stealthy and Precise Unmanned Aerial Vehicle Redirection with Real-Time Tracking and GPS~Spoofing}

\author{Haocheng Meng}
\affiliation{%
  \institution{Duke University}
  \city{Durham}
  \state{North Carolina}
  \country{USA}
}

\author{Shaocheng Luo}
\affiliation{%
  \institution{Duke University}
  \city{Durham}
  \state{North Carolina}
  \country{USA}
}

\author{Songqiao Xie}
\affiliation{%
  \institution{Duke University}
  \city{Durham}
  \state{North Carolina}
  \country{USA}
}

\author{Miroslav Pajic}
\affiliation{%
  \institution{Duke University}
  \city{Durham}
  \state{North Carolina}
  \country{USA}
}

\begin{abstract}
Redirecting unmanned aerial vehicles (UAVs) from their intended mission trajectories has been an active area of research. However, existing UAV redirection attacks lack reliability, precision, and covertness for a targeted diversion. They primarily rely on physical capture, communication hijacking, or sensor spoofing. Yet, physical interception is costly, offers only a single opportunity for success, and poses a high risk of collateral damage; network-based attacks demand deep technical expertise and access to encrypted communication channels; and sensor spoofing techniques typically fall short in achieving the accuracy and robustness required to steer a UAV toward a specified target. Consequently, we propose Phantom Navigator, a UAV redirection attack to mislead drones to a designated spoofing target, covertly and precisely. Our approach combines offline pre-redirection reachability analysis, which provides high-fidelity estimates of achievable redirect ranges, with an online closed-loop, stealthy execution layer that ensures successful redirection in practice. Based on this approach, we build a physical attack platform equipped with a LiDAR--camera detection, tracking, and spoofing stack that performs real-time identification, pose estimation, and computation of targeted spoofing signals to covertly and accurately redirect victim UAVs to a designated location. We demonstrate the effectiveness of our redirection methodology and the attack implementation in real-world case studies.
\end{abstract}


\maketitle

\section{Introduction}

Unmanned aerial vehicles (UAVs) have become integral to modern cyber-physical systems, 
in applications such as aerial mapping, package delivery, and infrastructure inspection. These platforms heavily rely  on onboard navigation, 
particularly the Global Positioning System (GPS), to localize, plan, and execute autonomous missions~\cite{miranda2022autonomous, patrik2019gnss}. Consequently, security of the navigation stack is critical to ensuring the success of 
UAVs' missions. 
Yet, 
recently it was shown that the integrity of the navigation systems 
can be compromised, 
forcing a UAV from the expected mission/trajectory (e.g.,~see~\cite{khan2021gps} and references therein).

\begin{figure}[!t]
    \centering
\includegraphics[width=0.892\columnwidth]{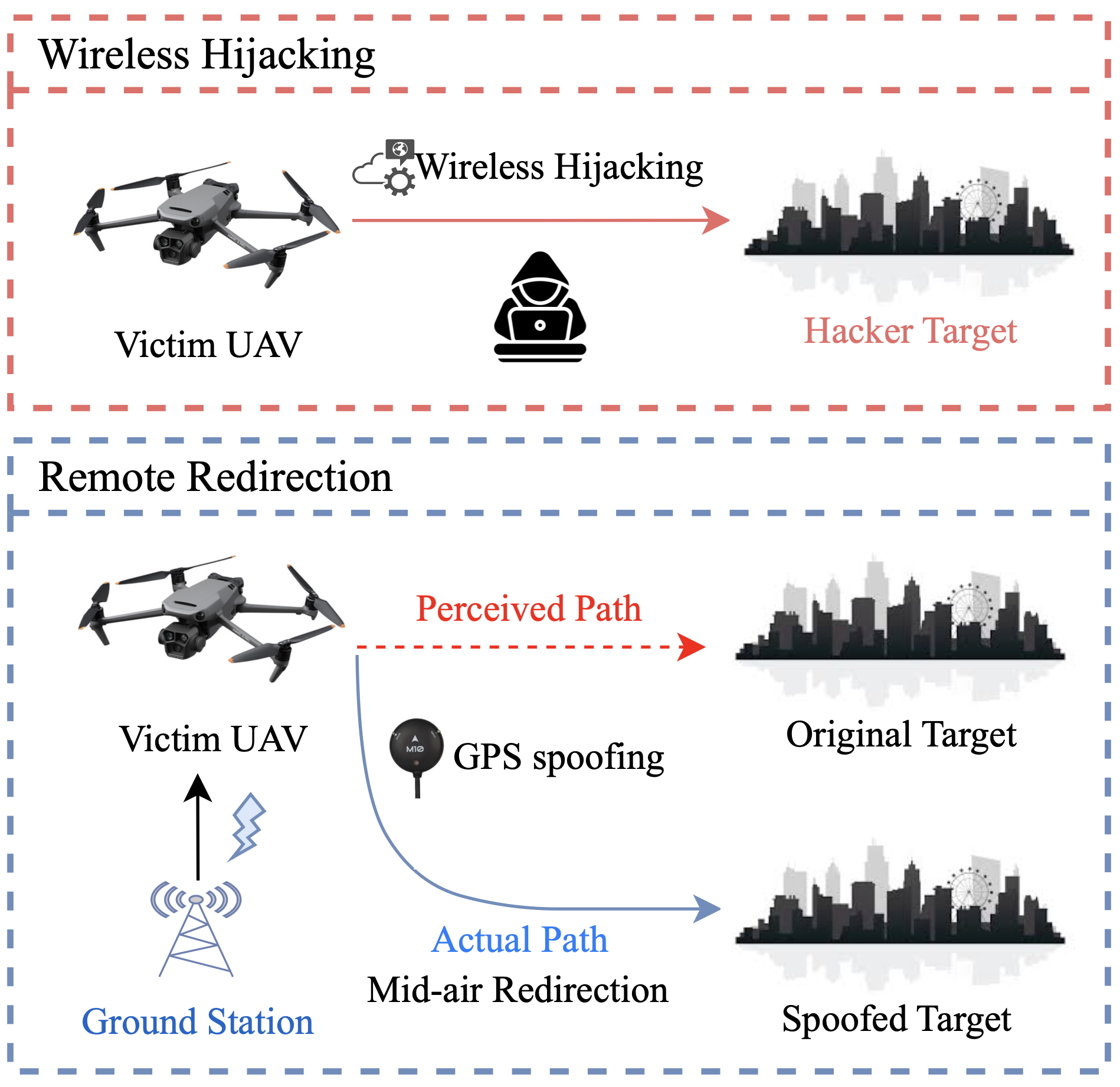}
    \caption{Existing UAV redirection attack approaches: ($i$)~Wireless hijacking; ($ii$) 
    Redirection using GPS spoofing.}
    \label{fig:redirection_methods}
\end{figure}

From the UAV navigation's perspective, 
two components are critical for mission integrity: 
($i$) the reference state, which encodes mission-level commands and trajectories, and ($ii$) the estimated state, reflecting the UAV’s (self) perceived position and orientation. Adversarial manipulation of either can redirect the vehicle from its intended trajectory. Correspondingly, as is depicted in Fig.~\ref{fig:redirection_methods}, existing attack methodologies can be broadly categorized into two classes: ($i$)~wireless hijacking and ($ii$)~sensor spoofing. 

Wireless hijacking seeks to compromise the communication or control interface, gaining unauthorized access to the navigation stack to alter mission commands directly~\cite{pratama2024behind, rubbestad2021hacking, 8658279}. In contrast, sensor spoofing perturbs the perception or sensor-fusion pipeline to induce erroneous state estimates, 
misleading the UAV’s 
control and decision-making~\cite{xu2023sok}. 
While 
former offers complete UAV control once successful, it is difficult to execute 
as it demands knowledge of proprietary protocols, cryptographic keys, or system vulnerabilities. 
Sensor spoofing, by comparison, presents a more practical and scalable attack surface that is cost-effective, hardware-agnostic, and physically deployable without prior access to the victim~UAVs.

Sensor spoofing attacks manipulate a target drone’s state estimation process by disrupting sensing and perception through various physical attack vectors~\cite{kim2024systematic}. Prior research has exposed vulnerabilities of UAV onboard sensors including GPS spoofing~\cite{noh2019tractor,sathaye2022experimental, chae2023commercial,chae2024reinforcement}, IMU acoustic injection~\cite{son2015rocking, tu2018injected, jeong2023rocking}, magnetic interference~\cite{jang2023paralyzing}, and visual deception against optical flow sensors~\cite{9444889, zhou2022doublestar}. 
%
Depending on the sensing modality, different attacks target distinct components of the UAV sensor fusion pipeline, and have different effects in diverting UAVs from their original trajectories. 
Inertial sensor attacks (e.g.,~\cite{son2015rocking, tu2018injected, jeong2023rocking}) corrupt the attitude estimation by injecting errors into body-frame angular rate measurements, causing severe degradation of the local inertial navigation system and immediate crashes. Although effective at destabilizing flight control, such attacks lack the precision required to steer the victim toward a specific location. Similarly, magnetic interference attacks affect the circuitry of onboard magnetometers and other sensors, leading to data corruption, signal saturation, and communication loss, making the attack outcome unpredictable, mostly resulting in crashes~\cite{jang2023paralyzing}.

Optical and vision-based attacks compromise the visual perception pipeline by injecting or distorting light signals, producing false or missing visual cues in the UAV’s environment 
\cite{9444889, zhou2022doublestar}; however, their impact is highly environment dependent and thus, difficult to control. In contrast, GPS spoofing attacks offer a uniquely controllable and targeted means of manipulation, by directly altering the localization and navigation 
estimates 
and enabling adversaries to systematically reroute victim drones toward designated~targets.

Nevertheless, the existing GPS spoofing attacks leave a question unanswered: \textit{Can a UAV be diverted, not only away from its course, but to an attacker-chosen destination without triggering anomaly detection?} UAVs, as most safety-critical systems, have anomaly detectors that revert system to safe operation in case of an anomaly. Therefore, stealthiness (i.e., covertness) is key to successful attacks because an overt, brutal force attack can be detected by UAV intrusion detection systems and the affected sensing modality could be filtered or removed, resulting in attack failure~\cite{8453641, zhong2021unmanned}. In practice, prior approaches fall short of achieving a targeted redirection attack with both the precision and stealthiness required for a truly controlled attack.

Leveraging spoofed GPS receivers, rule-based algorithms have been used to deviate UAVs from their paths~\cite{noh2019tractor,sathaye2022experimental}. However, these methods rely on open-loop control or human supervision, which makes them unreliable for accurate redirection to spoofing targets as the feedback to attack system is only dependent on human observations. 
Closed-loop approaches (i.e., where the inserted `spoofed' signal depends on the state of the UAV) have been explored for fixed-wing UAVs, aided by radars, to reroute the drones toward the target positions~\cite{chae2023commercial,chae2024reinforcement}. However, such attacks are easily detectable by anomaly/intrusion detection systems as they introduce out-of-distribution sensing errors, triggering statistical detectors (e.g., $\chi^2$ detector).  While~\cite{guo2019covert} proposes a stealthy spoofing method, it assumes complete knowledge of the victim UAV’s dynamics and parameters, which are often inaccessible and may vary with mission configurations -- thus, resulting in attacks being detected due to parameter changes that occur during every mission (e.g., battery level changes, wind, payload weight).

Consequently, prior research on GPS redirection attacks remains constrained in two aspects: (i) the inability to precisely control the spoofing outcome, and (ii) the lack of strategies to evade onboard anomaly detection mechanisms. These limitations significantly reduce the practicality, scalability, and reliability of existing GPS spoofing–based redirection attacks.

In this work, we address these limitations and show that it is possible to launch \emph{stealthy} GPS-spoofing attacks on UAVs that force the vehicles into the desired location. To achieve this, we introduce \emph{Phantom Navigator}, a real-time UAV redirection attack framework designed to precisely and covertly redirect victim drones to designated areas, regardless of their ongoing missions or onboard anomaly detection mechanisms. Building upon a formulation of GPS spoofing within UAV dynamics, we first conduct an efficient pre-attack redirection range analysis that estimates the achievable spoofing outcomes under black-box settings and varying levels of attacker knowledge. We then develop a lightweight, model-free redirection controller that computes GPS attack signals in real time based on the estimated victim pose and the spoofing target. 
To validate the framework, we implement an attacker platform prototype equipped with a real-time attack pipeline that performs identification and tracking through LiDAR–camera fusion, attack outcome estimation, and closed-loop spoofing signal generation. 

Our main contributions are summarized as follows.

\begin{itemize}
\item We introduce a novel UAV redirection attack featuring both offline attack range estimation and online closed-loop attack execution. Our framework provides both attack stealthiness and precision without assumption on the internal parameters or the employed sensor fusion pipeline. 

\item We build an experimental attacker platform that incorporates LiDAR–camera real-time onboard tracking and spoofing capabilities against victims equipped with GPS system to achieve redirection to desired target areas.

\item We evaluate our redirection attacks methodology and the attacker platform prototype on a victim drone in physical case studies, in both indoor controlled environments and outdoor real-world scenarios.
\end{itemize}

The paper is organized as follows. Sec.~\ref{sec:related_work} provides related work on UAV detection and tracking and the physical GPS attacks; Sec.~\ref{sec:threat_model} introduces the threat model, including attacker knowledge, capability, goal and implementation; Sec.~\ref{sec:methodology} presents Phantom Navigator methodology, with the redirection model, pre-attack redirection range estimation and in-attack controller design; Sec.~\ref{sec:experiments} demonstrates the physical design of Phantom Navigator and the real-world case studies both indoors and outdoors.

\section{Related Work}
\label{sec:related_work}

\subsection{UAV Detection and Tracking}
Accurate detection, identification, and tracking are essential prerequisites for reliable UAV redirection attacks. A variety of sensing modalities have been explored to detect UAVs across different ranges. Radar-based detection leverages RF reflections and offers the advantage of long-range operation {(e.g., up to $2~km$~\cite{de2018drone})} under diverse weather conditions~\cite{musa2019review,semkin2021drone,shin2016distributed,coluccia2020detection}. Conventional radar systems are effective for large aerial targets but often struggle with small drones due to their low speeds and weak radar signatures. To address this, 
mmWave radars have been employed, providing higher resolution and improved performance on smaller targets.

Vision-based detection has become a common practice with the rise of deep learning techniques~\cite{ge2021yolox,unlu2019deep,muller2017robust,seidaliyeva2020real,sharjeel2021real}. Camera systems can provide both detection and identification of drones, enabling classification and recognition. However, their performance significantly degrades under poor lighting, fog, or occlusion, making them less reliable in unstructured environments. LiDAR-based detection enables accurate three-dimensional localization and tracking, while being resilient to lighting conditions~\cite{dogru2022drone,hammer2018LiDAR,abir2023towards,hammer2019uav}. Similarly, acoustic detection methods have been studied to identify the unique noise signatures produced by UAV propellers, although these approaches face challenges in noisy urban environments~\cite{sedunov2019stevens,busset2015detection,fang2022drone,shi2020acoustic}. Recent state-of-the-art approaches integrate these modalities through multi-sensor fusion and leverage machine learning models to enhance robustness, reliability, and coverage~\cite{svanstrom2022drone,svanstrom2021dataset,svanstrom2021real,ding2023drone}.

\subsection{Physical GPS Take-over Attacks}

GPS spoofing manipulates vehicle localization by forging satellite signals, making the receiver compute false positions or time. 
As civilian GPS signals lack authentication, attackers can generate counterfeit signals using software-defined radios (SDRs) to mislead receivers into arbitrary trajectories~\cite{tippenhauer2011requirements}. A typical research setup (e.g.,~\cite{sathaye2022experimental}) employs an SDR-based GPS signal generator, an RF amplifier, and an omnidirectional transmit antenna, operated in a shielded anechoic chamber with a motion capture system used for ground truth tracking; the anechoic chambers are used because transmitting spoofed GPS signals over the air is illegal except in government-approved setups~\cite{murray2019legal_gnss_spoofing_wp}.

In general, GPS spoofing 
differentiates between overshadow (i.e., overpower) and take-over attacks~\cite{sathaye2022experimental}. {The overshadow attack transmits counterfeit GPS signals at a much higher power trying to bury legitimate signals, which is easy to detect due to the abrupt changes at the GPS receiver, resulting in a sudden loss of satellite lock.} 
To bypass the potential integrity checks at GPS receivers, with the take-over attacks, the spoofed signals are synchronized with legitimate ones first; the power is then gradually increased to induce false position estimates. Finally, this 
is followed by sophisticated post-takeover (spoofed) signal design to evade innovation checks in sensor fusion (e.g., with IMU).

\section{Threat Model}
\label{sec:threat_model}

In this work, we focus on GPS spoofing attacks that \emph{redirect} a UAV from its intended trajectory to an (attacker) selected target zone. Specifically, we consider the threat model 
where the attacker's objective is to \emph{precisely detour the UAV to a desired location, while remaining covert (i.e., stealthy) to \emph{a potential} onboard anomaly detection mechanism}. 
Since prior work fully demonstrated successful real-world GPS takeover attacks~\cite{sathaye2022experimental}, and we lack access to a 
shielded anechoic chamber or
government-authorized testing fields~\cite{murray2019legal_gnss_spoofing_wp}, we restrict our study to \emph{cyber-level GPS spoofing emulation}, rather than over-the-air signal transmission; in particular, we effectively insert the spoofed signals with characteristics described~in~\cite{sathaye2022experimental}.


Existing research primarily focuses on fixed-position ground-to-air tracking and spoofing~\cite{chae2023commercial, chae2024reinforcement}. In this work, we consider a stationary attacker model realized by a lightweight sensing and computation device that supports real-time victim tracking, spoofing signal generation, and closed-loop redirection. The proposed \emph{Phantom Navigator} therefore departs from heavy fixed ground-station assumptions and instead captures a more practical attacker setup with compact perception and computation, while retaining the key challenge of continuously steering the victim based on attacker-side observations.

\vspace{2pt}
\noindent\textbf{Attacker Knowledge and Capability.}
The adversary operates a lightweight stationary \emph{attacker platform} equipped with onboard perception and computation. This platform can: $(i)$ detect, identify, and track the victim UAV in real time using \emph{LiDAR--camera fusion}; $(ii)$ estimate the victim UAV's GPS noise characteristics; and $(iii)$ emulate GPS spoofing effects at the cyber level. Note that the GPS specifications of commercial UAVs are typically publicly available through manufacturer websites, user manuals, or datasheets (e.g.,~\cite{holybro-m10-gps}), from which the attacker can estimate the victim's GPS noise range. The adversary is \textbf{NOT} assumed to know the victim UAV's internal sensor-fusion parameters, communication protocols, or mission details beyond its GPS profile. However, any additional knowledge of the victim mission or operating region can further help the attacker select effective redirection targets.

\vspace{2pt}
\noindent\textbf{Attack Objective.} The goal is not to crash or disable the victim UAV, but to \emph{redirect it precisely and covertly} to a controlled area. As we consider \emph{black-box} attacks, where the attacker has no knowledge about the victim's autonomy stack (including which anomaly detector might be employed), it is required that the attack is stealthy to \emph{any} anomaly detection mechanisms.

\vspace{2pt}
\noindent\textbf{Attack Implementation.} The attack proceeds in three stages: $(i)$ Takeover Assumption: We assume the victim UAV’s GPS receiver has been taken over by spoofed signals (e.g., synchronized counterfeit signals). $(ii)$ Redirection Emulation: Our framework emulates the effects of spoofed GPS positions on the victim, generating controlled trajectory deviations that steer it away from its original mission path. $(iii)$ Termination: The spoofed trajectory is designed to precisely and covertly bring the victim to a designated target zone, avoiding fail-safe triggers or detection mechanisms.

\section{UAV Redirection Methodology}
\label{sec:methodology}

\begin{figure*}[!t]
    \centering
\includegraphics[width=0.98\textwidth]{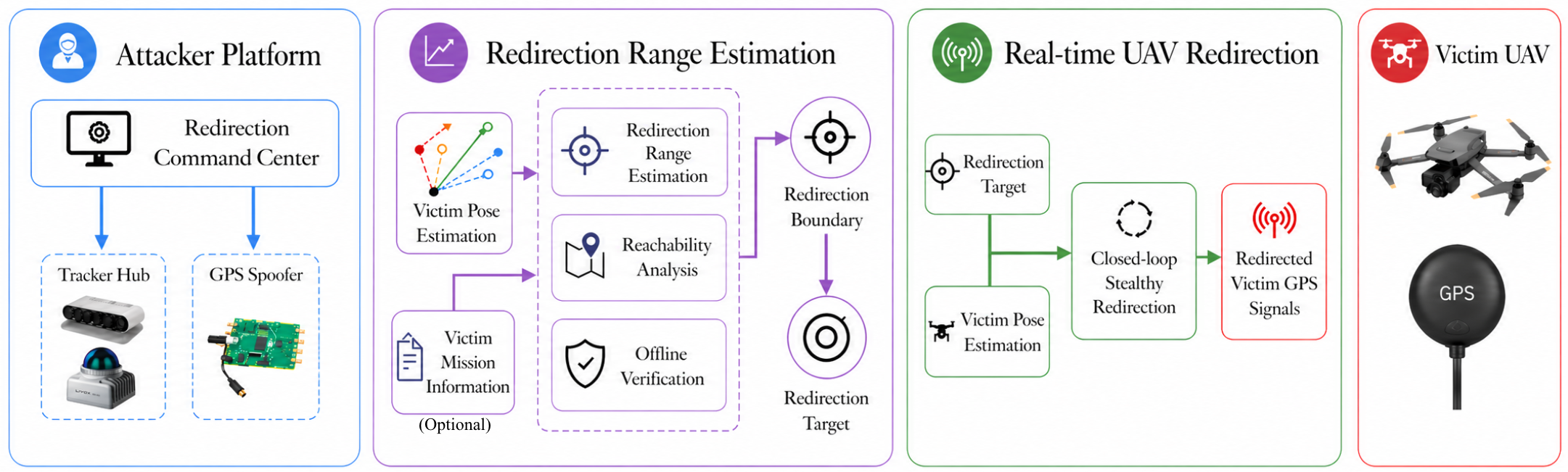}
    \caption{Overview of the proposed UAV redirection framework. The offline stage estimates feasible redirection range, while the online stage performs closed-loop stealthy GPS manipulation using real-time victim pose estimation.}
\label{fig:redirection_framework}
\end{figure*}

In summary, Phantom Navigator (shown in Fig.~\ref{fig:redirection_framework}) consists of three main components: $(i)$ attacker-side detection and tracking of the victim UAV; $(ii)$ redirection range estimation for identifying feasible target regions; and $(iii)$ closed-loop UAV redirection algorithm to achieve both \textit{precision} (being able to steer the UAV into the desired location) and \textit{stealthiness} against any potential anomaly detector. Specifically, our methodology focuses on estimating reliable victim redirection range before attack, and then performing real-time GPS-only redirection while preserving stealthiness.

This section presents the methodology as follows: We first formulate GPS-only stealthy redirection, then introduce reachability-based redirection range estimation under black-box settings and the improvement using different levels of mission knowledge, and finally describe the real-time closed-loop redirection controller.

\subsection{Stealthy Redirection Formulations}
\label{subsec:redirection_model}

In this work, a \emph{successful stealthy redirection} is defined as redirecting the victim UAV while remaining undetected by anomaly detectors that assess whether the observed system evolution is nominal. Since any practical detector must tolerate a nonzero false-alarm rate to remain robust to sensor noise, modeling inaccuracies, and environmental disturbances, an attacker can exploit this inherent tolerance by ensuring that the victim's believed navigation dynamics remain statistically close to attack-free operation. In particular, by perturbing only the GPS readings within the nominal GPS uncertainty envelope, the attacker can induce a biased trajectory whose effect accumulates over time into controlled redirection. This intuition is consistent with existing research on black-box stealthy GPS attack analysis~\cite{10885819}. Accordingly, we formulate stealthy UAV redirection by answering three questions: \textit{$(i)$ How redirection is achieved by manipulating GPS only; $(ii)$ Why this GPS manipulation is stealthy; $(iii)$ How this stealthiness is achieved in practice.}

\subsubsection{GPS-Only Redirection Mechanism}

In standard UAV navigation stacks, GPS provides direct measurements of earth-frame position and velocity, which are fused with IMU readings to produce the navigation state used by the autopilot. Importantly, GPS (or other GNSS) is typically the only onboard sensor that provides absolute global positioning information, making it a unique attack surface because this global reference cannot be directly cross-validated by other standard onboard sensors. Even if alternative signals of opportunity (SOPs) provide contradictory positioning cues, the UAV cannot readily determine whether GPS or the auxiliary signal is at fault. This \textit{ambiguity in trust assignment} makes stealthy attacks on the global reference particularly challenging to detect.

Under a GPS takeover attack, the adversary perturbs only the GPS channel by adding spoofing offsets to the reported position and velocity
~\cite{noh2019tractor,sathaye2022experimental,chae2023commercial,chae2024reinforcement}. As a result, the victim's sensor-fusion module reconstructs a corrupted but internally consistent navigation state from spoofed GPS and nominal IMU measurements, and the autopilot closes the loop on this biased state. \textit{Therefore, manipulating GPS alone is sufficient to alter the victim's believed motion and induce controlled redirection, without attacking other sensors.}

To capture this effect, we isolate the translational component of the  motion and use a linearized discrete-time state-space model:
\begin{equation}
\mathbf{x}_{k} = A\,\mathbf{x}_{k-1} + B\,\mathbf{u}_{k-1} + \mathbf{w}_{k-1},
\quad
\mathbf{y}_k = C\,\mathbf{x}_{k-1} + \mathbf{v}_{k-1},
\label{eq:redirection_dynamics}
\end{equation}
where $\mathbf{x}_k$ denotes the true translational state, $\mathbf{u}_k$ is the equivalent spoofing-induced redirection input, and $\mathbf{w}_k,\mathbf{v}_k$ capture process and measurement noise. In particular, we consider
\begin{equation}
\mathbf{x}_k =
\begin{bmatrix}
\mathbf{p}_k,\mathbf{v}_k,\mathbf{s}_{p,k},\mathbf{s}_{v,k}
\end{bmatrix},
\quad
\mathbf{u}_k=\mathbf{s}_{a,k},
\label{eq:redirection_state_control}
\end{equation}
where $\mathbf{p}_k,\mathbf{v}_k$ are the true position and velocity, and $\mathbf{s}_{p,k},\mathbf{s}_{v,k}$ denote the spoofing-induced offsets in GPS-reported motion. This reduced model captures how persistent GPS perturbations produce accumulated translational drift over time; the detailed UAV modeling and GPS spoofing formulations are provided in App.~\ref{app:uav_physical_dynamics} and App.~\ref{app:gps_spoofing_math}.

\subsubsection{Stealthiness Guarantees for Redirection}

Under GPS spoofing, the victim UAV closes the loop not on its true translational state $\mathbf{x}_k$, but on a \emph{believed} state $\tilde{\mathbf{x}}_k$ reconstructed from spoofed navigation measurements. Let $\pi(\cdot)$ denote the victim autopilot control law. The believed dynamics evolve as
\begin{equation}
\tilde{\mathbf{x}}_{k}
=
A\,\tilde{\mathbf{x}}_{k-1}
+
B\,\pi(\tilde{\mathbf{x}}_{k-1}),
\label{eq:believed_dynamics}
\end{equation}
while the true dynamics satisfy
\begin{equation}
\mathbf{x}_{k}
=
A\,\mathbf{x}_{k-1}
+
B\!\left(\pi(\tilde{\mathbf{x}}_{k-1}) + \mathbf{u}_{k-1}\right),
\label{eq:true_dynamics_with_belief}
\end{equation}
where $\mathbf{u}_{k-1}$ denotes the equivalent spoofing-induced redirection input. Defining the deviation
\begin{equation}
\mathbf{d}_k = \mathbf{x}_k - \tilde{\mathbf{x}}_k,
\end{equation}
we obtain
\begin{equation}
\mathbf{d}_{k}
=
A\,\mathbf{d}_{k-1}
+
B\,\mathbf{u}_{k-1},
\qquad
\mathbf{d}_0 = \mathbf{0}.
\label{eq:deviation_dynamics}
\end{equation}
Hence, the separation between the true and believed trajectories is governed by \textit{a linear deviation dynamics} driven by the spoofing input. This allows for gradual accumulation of controlled displacement while preserving a physically consistent believed motion.

To formalize stealthiness, let $\mathbf{z}_{0:k}^0$ and $\mathbf{z}_{0:k}^a$ denote the detector-visible observation sequences under nominal and attacked execution, respectively. We define the redirection to be $\epsilon$-stealthy if
\begin{equation}
D_{\mathrm{KL}}\!\left(
\mathbb{P}(\mathbf{z}_{0:k}^a)
\;\middle\|\;
\mathbb{P}(\mathbf{z}_{0:k}^0)
\right) \le \epsilon,
\qquad \forall k,
\label{eq:epsilon_stealthiness_obs}
\end{equation}
where $D_{\mathrm{KL}}$ is the Kullback--Leibler divergence and $\epsilon$ is the maximum admissible discrepancy between the attacked and nominal observation distributions. This condition captures detector-agnostic stealthiness: if the attacked observation process remains statistically close to the nominal one, no anomaly detector can reliably distinguish the attack beyond its tolerated false-alarm behavior~\cite{10885819}.

The key connection between Eqs.~\ref{eq:deviation_dynamics} and~\ref{eq:epsilon_stealthiness_obs} is that the detector does not observe the spoofing input directly; instead, it only observes the mismatch induced by the state deviation $\mathbf{d}_k$. Therefore, controlling the linear deviation dynamics directly limits the discrepancy between $\mathbf{z}_{0:k}^a$ and $\mathbf{z}_{0:k}^0$, thereby bounding the statistical footprint of the attack on detector-visible observations. \textit{As a result, if the spoofing input is chosen such that the induced deviation remains within the nominal GPS uncertainty envelope, the KL divergence stays bounded and the redirection retains a strong stealthiness guarantee.}

\subsubsection{Practical Stealthiness Calibration}

While Eq.~\ref{eq:epsilon_stealthiness_obs} provides a detector-agnostic formulation of stealthiness, practical attack synthesis requires an operational condition that can be directly enforced online. To this end, we translate the statistical stealthiness objective into a bound on the spoofing input magnitude. This calibration links the theoretical KL-based guarantee to a measurable attack parameter that can be tuned from victim's GPS noise profile.

\vspace{2pt}
\noindent\textbf{Realistic bound on spoofing input.}
In practice, we enforce stealthiness through a bounded spoofing-input constraint:
\begin{equation}
\|\mathbf{u}_k\| \le \lambda_a, \qquad \mathbf{u}_k \in \mathcal{A},
\label{eq:stealthiness_input_bound}
\end{equation}
where $\lambda_a$ denotes the maximum admissible spoofing-induced acceleration. This bound serves as a practical surrogate for the KL-based stealthiness condition in Eq.~\ref{eq:epsilon_stealthiness_obs}: bounding $\|\mathbf{u}_k\|$ controls the deviation dynamics in Eq.~\ref{eq:deviation_dynamics}, which in turn limits the observable mismatch between attacked and nominal executions. \textit{Importantly, $\lambda_a$ depends only on the victim's GPS noise profile: the admissible spoofing magnitude must remain masked by the nominal GPS uncertainty envelope, so noisier GPS measurements allow larger stealthy perturbations.}

\vspace{2pt}
\noindent\textbf{Estimating the spoofing bound $\lambda_a$.}
To estimate $\lambda_a$, we calibrate the largest spoofing magnitude whose detector response remains statistically indistinguishable from nominal operation. Let $\alpha$ denote the false-alarm rate under benign conditions. For a fixed spoofing magnitude $\lambda$, we run $N$ Monte Carlo trials using  the identified GPS noise profile $\boldsymbol{\Sigma}_{\mathrm{gps}}$. Each trial records whether the detector raises at least one alarm over the evaluation horizon. From these trials, we compute a $(1-\delta)$ confidence interval for the induced detection probability using the Clopper--Pearson method~\cite{clopper1934confidence}. We then define
\begin{equation}
\lambda_a(\delta;\boldsymbol{\Sigma}_{\mathrm{gps}})
\triangleq
\sup\big\{\lambda\ge 0:\ \alpha \in
\mathrm{CI}_{1-\delta}(\lambda;\boldsymbol{\Sigma}_{\mathrm{gps}})\big\},
\end{equation}
i.e., the largest spoofing bound whose detection probability remains statistically consistent with the nominal false-alarm rate. \textit{Thus, $\lambda_a$ provides an implementation-ready approximation of the maximum GPS-only spoofing strength that preserves stealthy redirection.}

\subsection{Redirection Range Estimation}
\label{subsec:pre_redirection_reach}
Before launching the attack, the attacker must estimate the effective redirection attack range, i.e., \textit{the farthest range to which the victim can be reliably deviated}. This range defines the feasible redirection region under stealthy spoofing inputs and enables the attacker to select targets that are both achievable and covert. In our framework, we estimate this range through reachability analysis of the redirection model under the calibrated spoofing bound $\lambda_a$.

\subsubsection{Reachability of Redirection System}

Given the redirection dynamics in Eq.~\ref{eq:redirection_dynamics}, we characterize the attacker's achievable redirection range through the reachable set of the system. At time step $k$, the reachable set is defined as
\begin{equation}
\mathcal{R}_k = \left\{ \mathbf{x}_k \,\middle|\, 
\mathbf{x}_{i+1} = A \mathbf{x}_i + B \mathbf{u}_i,\ 
\mathbf{u}_i \in \mathcal{U},\ i = 0, \dots, k-1 \right\},
\label{eq:reachable_set}
\end{equation}
where $\mathcal{R}_k$ contains all states that can be reached from the initial condition under admissible spoofing inputs. The input set is determined by the stealthiness-constrained spoofing bound $\lambda_a$:
\begin{equation}
\mathcal{U} = \left\{ \mathbf{u} \in \mathbb{R}^m \,\middle|\, \|\mathbf{u}\| \le \lambda_a \right\}.
\end{equation}

Since the attack objective is to redirect the victim to a desired physical location, we are primarily interested in the position component of the reachable set. We therefore project $\mathcal{R}_k$ onto the position states and obtain
\begin{equation}
\mathcal{P}_k = \left\{ \mathbf{p}_k \in \mathbb{R}^3 \,\middle|\, \exists\, \mathbf{x}_k \in \mathcal{R}_k \right\},
\label{eq:projected_position_set}
\end{equation}
where $\mathcal{P}_k$ represents the set of all victim positions that can be reached by time $k$ under stealthy redirection. \textit{Thus, $\mathcal{P}_k$ directly characterizes the feasible redirection region where the victim can be reliably steered at a given time.}

\subsubsection{Computational Solution to Reachability Analysis}

Direct computation of the exact reachable set can become expensive as the horizon grows. We therefore adopt a lightweight convex-hull approximation to efficiently estimate the stealthiness-constrained reachable set of the redirection system.

Given the discrete-time LTI dynamics in Eq.~\ref{eq:redirection_dynamics}, a compact convex input set $\mathcal{U}$, and initial condition $\mathbf{x}_0$, the reachable set at time $k$ can be approximated as
\begin{equation}
\mathcal{R}_k \approx A^k \mathbf{x}_0 \ \oplus\ 
\mathrm{conv}\left( \left\{ \sum_{i=0}^{k-1} A^{k-1-i} B\,\mathbf{u}_i^{(j)} \ \middle| \ \mathbf{u}_i^{(j)} \in \mathcal{U} \right\}_{j=1}^N \right),
\label{eq:convex_hull_full}
\end{equation}
where $\oplus$ denotes the Minkowski sum, $\mathrm{conv}(\cdot)$ denotes the convex hull, and $N$ is the number of sampled input sequences. Intuitively, each sampled sequence generates one candidate redirection path, and the convex hull of them provides an efficient outer approximation of the states reachable under stealthy spoofing inputs.

As $N$ increases, the approximation becomes tighter and better captures the feasible region. \textit{This convex-hull formulation provides a computationally efficient way to estimate the achievable redirection range without repeatedly solving the full nonlinear UAV dynamics.}

\subsubsection{Partial Information Refinement in Redirection}

The redirection attack fundamentally operates in the \emph{black-box} setting: the attacker has no knowledge of the victim's mission beyond observable motion and the GPS profile. In this baseline case, the reachable set in Eq.~\ref{eq:convex_hull_full} provides a \textit{conservative} estimate of the successful redirection range under stealthy attacks. \textit{Thus, victim mission knowledge is NOT required for the attack; instead, it expands the range over which the attacker can confidently achieve successful redirection.}

When partial mission information is available, the attacker can refine the range estimate without performing the actual redirection. Under GPS spoofing, the victim UAV closes the loop on its \emph{believed} state $\tilde{\mathbf{x}}_k$ while the true state $\mathbf{x}_k$ is manipulated by the redirection algorithm, as described in Eqs.~\ref{eq:believed_dynamics} and~\ref{eq:true_dynamics_with_belief}. We encode available mission knowledge as a belief-consistency constraint
\begin{equation}
\tilde{\mathbf{x}}_k \in \tilde{\mathcal{X}}^{\text{mission}}_k, \qquad \forall k,
\label{eq:stealthiness_constraint}
\end{equation}
where $\tilde{\mathcal{X}}^{\text{mission}}_k$ denotes the set of believed states consistent with the revealed mission information. The refined reachable set is
\begin{equation}
\mathcal{R}_k^{\mathrm{part}} = 
\left\{ \mathbf{x}_k \ \middle|\ 
\exists\ \tilde{\mathbf{x}}_i \in \tilde{\mathcal{X}}^{\mathrm{mission}}_i,\ 
\mathbf{u}_i \in \mathcal{U},\ 
\forall i < k 
\right\},
\label{eq:partial_info_reachability}
\end{equation}
which restricts the analysis to trajectories that remain consistent with both stealthy attacks and available mission knowledge. This refinement removes implausible worst-case mission evolutions from the black-box estimate, allowing the attacker to certify a less conservative and potentially larger guaranteed redirection range.

Specifically, we consider four knowledge levels: $(0)$ \emph{Black-box}, where no mission information is available; $(1)$ \emph{Destination}, where only the final target is known; $(2)$ \emph{Mission trajectory}, where the geometric route is known but not its precise timing; and $(3)$ \emph{Motion planning}, where both the route and timing are known. As the attacker gains more mission information, the range estimate becomes less conservative, enabling the attacker to select farther targets with higher confidence. A detailed summary is given in Table~\ref{tab:mission_knowledge_levels}.

\begin{table*}[!t]
\centering
\renewcommand{\arraystretch}{1.2}
\caption{Mission knowledge levels and their impact on redirection range estimation.}
\begin{tabular}{p{0.15\textwidth} p{0.20\textwidth} p{0.20\textwidth} p{0.35\textwidth}}
\hline\hline
\textbf{Knowledge} & \textbf{Mission Assumption} & \textbf{Worst-case Prediction} & \textbf{Effect on Redirection Range} \\
\hline\hline
0--Black-box &
No victim mission information is known. &
Victim may follow any feasible nominal evolution. &
Produces the most conservative baseline estimate; no prior knowledge is required for attack. \\
\hline
1--Destination &
Only the final destination is known. &
Victim heads directly toward the destination at maximum speed. &
Removes missions inconsistent with the known destination, expanding the range where redirection can be achieved with higher confidence. \\
\hline
2--Mission trajectory &
The geometric route is known, but not its precise timing. &
Victim traverses the known route at maximum speed. &
Further removes path-inconsistent evolutions, yielding a less conservative and more informative range estimate. \\
\hline
3--Motion planning &
Both the route and timing are known. &
Worst-case and nominal evolution coincide. &
Provides the least conservative estimate, with the smallest uncertainty in achievable rerouting. \\
\hline\hline
\end{tabular}
\label{tab:mission_knowledge_levels}
\end{table*}

\begin{figure*}[!t]
    \centering
\includegraphics[width=0.98\textwidth]{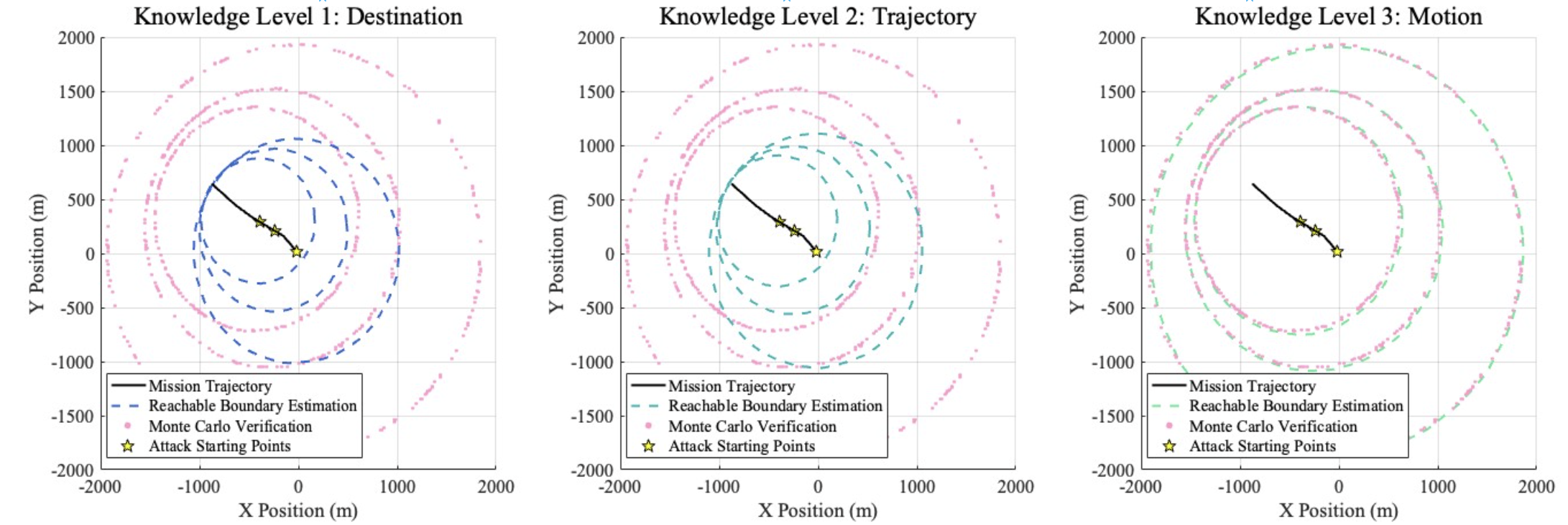}
    \caption{Redirection range estimation compared with Monte-Carlo verification under different knowledge levels. Left to right: destination-only, mission trajectory and motion planning, with increased overlap with Monte-Carlo results.}
    \label{fig:redirection_range_eg}
\end{figure*}

\subsubsection{Redirection Range Verification}

To validate the estimated redirection range, we perform offline Monte-Carlo simulations without executing the attack on a physical UAV. At selected timestamps during the victim mission, we first compute the redirection range using the convex-hull reachability approximation, and then run Monte-Carlo rollouts from the same initial conditions with simulated spoofing targets. These rollouts provide an empirical outer boundary of achievable redirection outcomes for comparison.

An illustrative example is shown in Fig.~\ref{fig:redirection_range_eg}. We load the quadcopter dynamics (details in App.~\ref{app:uav_physical_dynamics})  and the redirection model into MATLAB/Simulink (details in App.~\ref{app:simulation_setup}) and evaluate three mission stages during a 400\,s waypoint mission. Since the black-box case (Level~0) already provides the baseline range estimate without requiring any mission knowledge, we focus on verifying Levels~1--3, where partial mission information refines the predicted range. For each of these levels, we adopt the corresponding worst-case assumptions in Table~\ref{tab:mission_knowledge_levels}, yielding conservative redirection range estimates. As shown in Fig.~\ref{fig:redirection_range_eg}, the Monte-Carlo results form a broader empirical boundary, while the reachability-based estimates remain tighter and conservative, with increasing precision as more mission information becomes available. \textit{Thus, the verification confirms that our method identifies a guaranteed redirection region: any target chosen inside the estimated boundary can be reliably reached under the assumed stealthiness and mission-consistency constraints.}

\subsection{Real-time Closed-loop UAV Redirection}

After estimating the feasible redirection range, the attacker needs an online mechanism to steer the victim UAV toward a chosen target. We therefore develop a real-time closed-loop redirection algorithm that updates spoofed GPS position and velocity signals based on the tracked victim state and the desired target. This controller is designed to achieve target-oriented redirection while remaining within the stealthiness constraints derived earlier. We next present the controller design and verify its effectiveness in simulation.

\subsubsection{Redirection Controller Design}

\begin{algorithm}[!t]
\caption{Stealthiness-Guaranteed Closed-loop Redirection}
\label{algo:stealthy_redirection}
\begin{algorithmic}[1]
\STATE \textbf{Input:} redirection target $p_d$, desired terminal velocity $v_d$, spoofing bound $\lambda_a$
\STATE \textbf{Output:} spoofed GPS signals $Y^a_{p,\mathrm{GPS}}, Y^a_{v,\mathrm{GPS}}$
\STATE \textbf{Initialize:} spoofing states $p_{s,0}=0$, $v_{s,0}=0$

\WHILE{GPS takeover remains effective}
    \STATE Update victim state estimates $(\hat{p}_k,\hat{v}_k)$ from attacker-side tracker
    \STATE Compute target-tracking errors $(\Delta p_k,\Delta v_k)$
    \STATE Generate target-oriented spoofing acceleration candidate $a_{s,k}=f_s(\Delta p_k,\Delta v_k)$
    \STATE Enforce stealthiness by clipping $a_{s,k}$ to satisfy $\|a_{s,k}\|\leq \lambda_a$
    \STATE Update spoofing states $(v_{s,k},p_{s,k})$ by discrete integration
    \STATE Construct and inject spoofed GPS outputs $(Y^a_{p,\mathrm{GPS}},Y^a_{v,\mathrm{GPS}})$
\ENDWHILE
\end{algorithmic}
\end{algorithm}

\begin{figure*}[!t]
    \centering
\includegraphics[width=0.97\textwidth]{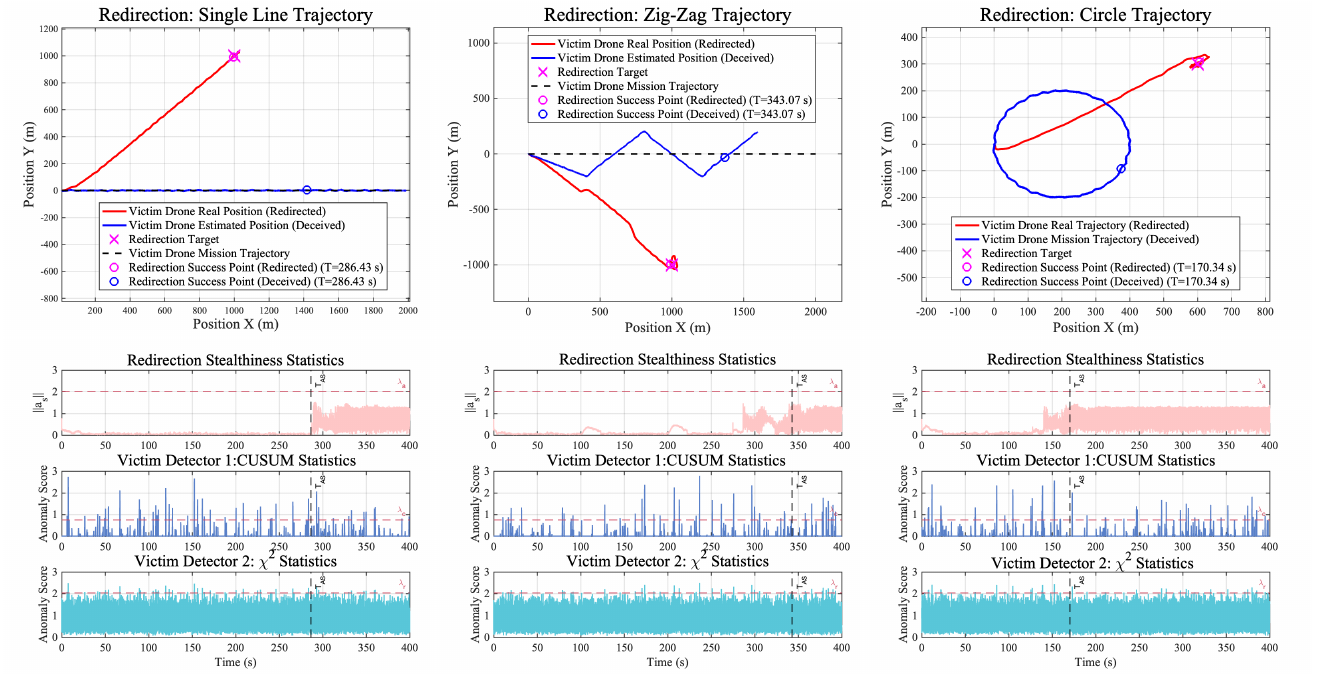}
    \caption{Redirection control results and stealthiness statistics of three different victim mission trajectories. Left to right: single line, zig-zag and circle trajectory. The statistics include redirection acceleration norm, CUSUM and $\chi^2$ anomaly detector scores.}
    \label{fig:redirection_control_results}
\end{figure*}

To remain compatible with the black-box GPS-only threat model, we design a lightweight feedback controller that operates directly in the GPS measurement space, without requiring access to the victim's internal control stack. Algorithm~\ref{algo:stealthy_redirection} summarizes this execution flow.

The controller takes as input a desired redirection target $p_d$ and the calibrated spoofing bound $\lambda_a$. At each control step, the attacker updates the victim's current position and velocity estimates, denoted by $\hat{p}_k$ and $\hat{v}_k$, from the onboard tracker, and computes the position and velocity errors relative to the target:
\begin{equation}
\Delta p_k = p_d - \hat{p}_k, \qquad
\Delta v_k = v_d - \hat{v}_k .
\end{equation}
These errors are mapped to a target-oriented spoofing acceleration
\begin{equation}
a_{s,k} = f_s(\Delta p_k,\Delta v_k),
\end{equation}
where $f_s(\cdot)$ is a lightweight proportional-derivative feedback law. The key design choice is not the feedback law itself, but the explicit enforcement of stealthiness in the spoofing space.

To preserve stealthiness, the spoofing acceleration is clipped to the admissible bound $\lambda_a$ while preserving its direction:
\begin{equation}
a_{s,k} =
\begin{cases}
\dfrac{\lambda_a}{\|a_{s,k}\|} a_{s,k}, & \text{if } \|a_{s,k}\| > \lambda_a, \\[4pt]
a_{s,k}, & \text{otherwise}.
\end{cases}
\end{equation}
The clipped acceleration is then integrated to update the spoofed velocity and position states,
\begin{equation}
v_{s,k} = v_{s,k-1} + a_{s,k}\Delta T,
\qquad
p_{s,k} = p_{s,k-1} + v_{s,k}\Delta T,
\end{equation}
which are finally injected into the victim's GPS outputs:
\begin{equation}
Y^a_{p,\mathrm{GPS}} = Y_{p,\mathrm{GPS}} - p_{s,k}, \qquad
Y^a_{v,\mathrm{GPS}} = Y_{v,\mathrm{GPS}} - v_{s,k}.
\end{equation}

{Thus, the controller realizes target-oriented UAV redirection by continuously biasing the victim's GPS position and velocity, while the stealthiness bound $\lambda_a$ ensures that the injected perturbations remain within the admissible GPS uncertainty envelope.}

\subsubsection{Redirection Control Verification}

We verify the proposed closed-loop redirection controller through simulation on three representative victim mission trajectories: single-line, zig-zag, and circular motion. For each case, the attacker applies Algorithm~\ref{algo:stealthy_redirection} to steer the victim toward a designated redirection target while enforcing the calibrated spoofing bound $\lambda_a$. As shown in Fig.~\ref{fig:redirection_control_results}, the controller drives the victim's \emph{true} trajectory toward the target, while the victim's \emph{believed} trajectory remains close to its nominal mission path. {This confirms the key objective of the controller: achieving target-oriented redirection in the physical world while preserving nominal-looking GPS evolution from the victim's perspective.}

\vspace{2pt}
\noindent\textbf{Controlled redirection performance.}
Across all three mission types, the controller achieves effective target-oriented rerouting despite the very different nominal motion patterns. In the single-line case, the victim is redirected away from its straight path toward a distant target; in the zig-zag case, the controller induces a substantial downward deviation while maintaining the victim's believed motion near the original path; and in the circular case, the victim is steered away from the orbit toward the designated destination. These results show that the controller is not tied to a specific nominal trajectory, but can adapt to different mission geometries using only attacker-side tracking feedback and GPS-space spoofing.

\vspace{2pt}
\noindent\textbf{Stealthiness against detectors.}
Fig.~\ref{fig:redirection_control_results} also reports the spoofing acceleration  with the victim-side CUSUM and $\chi^2$ detector statistics. In all three cases, the spoofing signal remains clipped by the stealthiness bound, and both detector statistics stay near nominal operating region without sustained threshold crossings during redirection. This shows that the controller not only achieves controllable deviation, but also keeps the induced observation mismatch within the admissible stealthiness envelope. Overall, the simulation results verify that the proposed controller can realize effective closed-loop UAV redirection while maintaining strong practical covertness.
\section{Real-world Redirection Study}
\label{sec:experiments}

\begin{figure*}[!t]
    \centering
\includegraphics[width=0.95\textwidth]{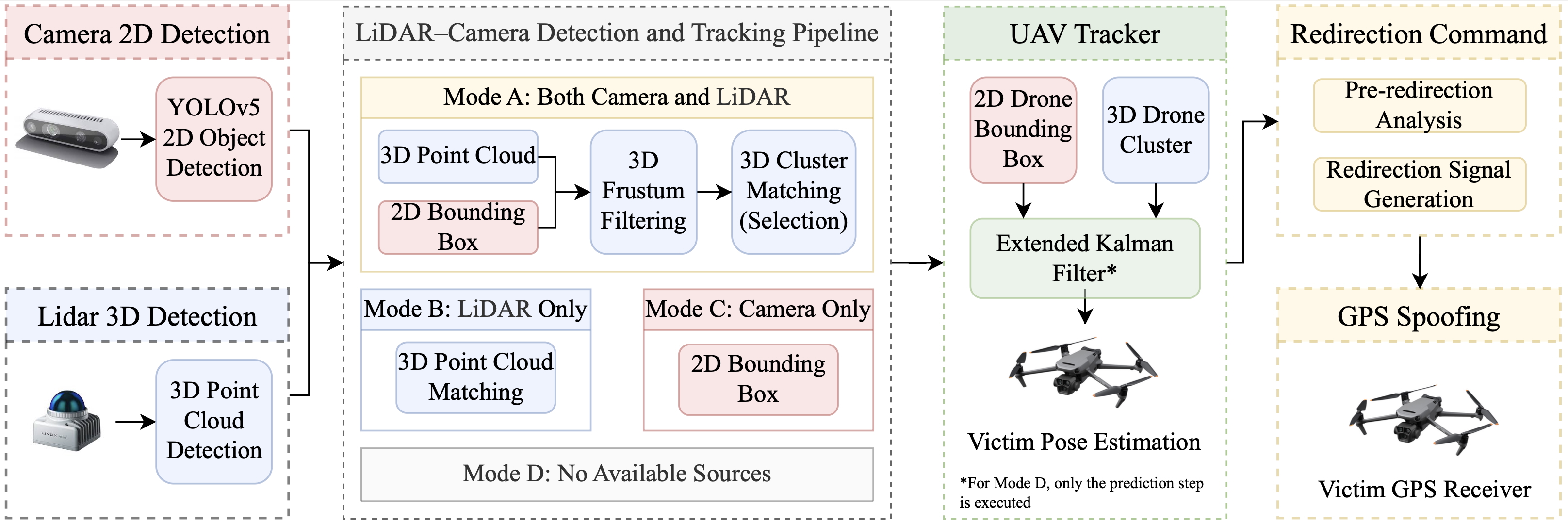}
    \caption{Full-stack attacker platform for stealthy UAV redirection, with LiDAR-camera detection, tracking and spoofing.}
    \label{fig:redirection_implementation}
\end{figure*}

\begin{figure*}[!t]
    \centering
\includegraphics[width=\textwidth]{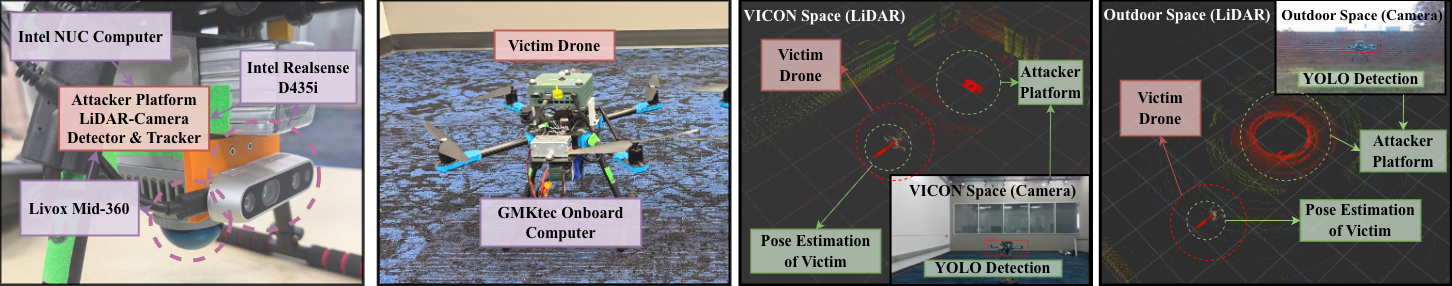}
    \caption{Experimental setup and sensing pipeline. Left: attacker-side tracking platform equipped with an Intel NUC, RealSense D435i, and Livox Mid-360. Middle-left: victim UAV with onboard mission computer. Middle-right and right: LiDAR--camera victim detection and pose estimation in indoor VICON and outdoor GPS-enabled environments.}
    \label{fig:exp_setup}
\end{figure*}

To implement the complete UAV redirection framework, we design an attacker platform with an onboard redirection station with a detection and tracking sensor hub. The full-stack implementation for this platform is illustrated in Fig.~\ref{fig:redirection_implementation}, targeting the victims. Deployed on the onboard computer of attacker platform, a LiDAR–camera tracking hub captures the $2D$ image and $3D$ point cloud of any potential victim UAV, and conducts both \textit{object detection} and \textit{pose estimation}. Then, a spoofing module computes the desired GPS spoofing signal based on the spoofing target and the estimated pose of the victim. Utilizing this platform, we design real-world redirection case studies and demonstrate the effectiveness of the framework. Next, we provide the implementation details and the evaluation of the real-world case studies, both indoors and outdoors.

\subsection{Experimental Design}

\subsubsection{Attacker Platform and Victim UAV} 

We built the attacker platform on a Holybro X500 V2 airframe and an Intel NUC for onboard processing. To enable redirection, we equipped it with an Intel RealSense D435i camera and a Livox Mid-360 LiDAR, mounted in forward and downward-facing orientations, respectively. While this tracking hub remained \textit{stationary during all experiments} and operated as a fixed tracking platform, its airframe-based installation lays the foundation for future moving-attacker deployments. The victim UAV used a Pixhawk 6X flight controller on Holybro X500 V2 airframe, and relied on a lightweight GMKtec onboard computer for mission execution. Except for the GPS noise profile, all other victim-side parameters are unknown to the attacker, ensuring a black-box attack setting. This setup is illustrated in Fig.~\ref{fig:exp_setup}.

\subsubsection{LiDAR–Camera Detection and Tracking Pipeline} 
\label{subsec:visual_lidar_pipeline}

To both identify and track the victim drone, we leverage a LiDAR–camera combined detection and tracking pipeline as shown in Fig.~\ref{fig:redirection_implementation}, which features 4 different modes according to the availability of the tracking sources in real-time. Camera image is processed by YOLOv5 real-time $2D$ object detection to provide bounding box; LiDAR $3D$ point cloud is filtered, clustered and matched in the frustum defined by the camera bounding box or processed alone.
Then, a Kalman-filter based tracker is implemented to estimate the real-time position and velocity of the drone. By accounting for temporary detection and tracking losses due to obstacles or weather changes in the 4 different modes, our tracker provides a consistent and reliable pose estimation of the victim drone.

\subsubsection{GPS Spoofing Emulation}

Our study assumes a successful GPS takeover and focuses on the post-takeover redirection phase. As physical GPS spoofing is outside the scope of this work, we emulate the spoofing mechanism in software and inject false navigation data directly into the victim UAV. The spoofing signals are computed online by the attacker tracking platform using the victim's estimated pose and the designated redirection target, and are then transmitted to the victim for false data injection.

The \textit{difference between indoor and outdoor experiments} lies in the navigation source used for emulation. Indoors, GPS measurements are emulated from VICON; outdoors, the attack is applied to the victim's North-East-Down (NED) navigation frame derived from real GPS measurements in Latitude-Longitude-Altitude (LLA) coordinates. This enables consistent evaluation of the proposed redirection framework across both indoor and outdoor settings.

\subsubsection{Computation Overhead}

\begin{figure}[!t]
    \centering
\includegraphics[width=\columnwidth]{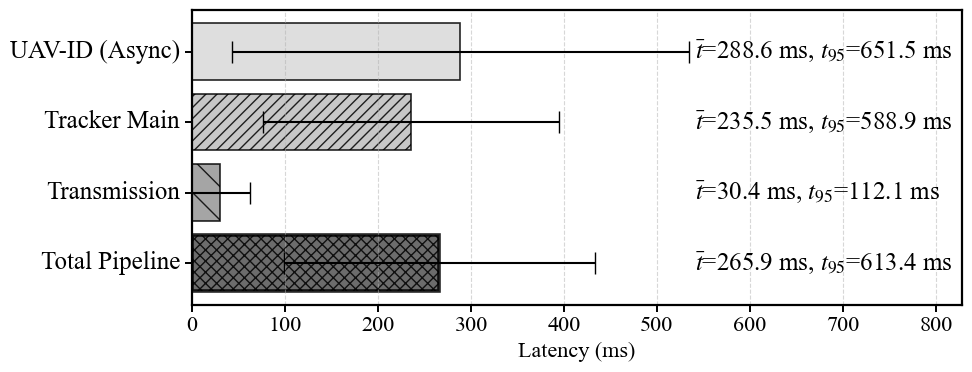}
    \caption{Physical pipeline latency of tracker, attacker to victim transmission and asynchronous UAV identification.}
    \label{fig:comp_overhead}
\end{figure}

The computation is measured on the Intel NUC onboard computer, where the tracker uses $73.9\%$ of the CPU and $4.7\%$ of the RAM, and is GPU-free. Fig.~\ref{fig:comp_overhead} shows the latency breakdown: The end-to-end tracking and attack path has an average latency of 265.9\,ms and a 95th-percentile latency of 613.4\,ms, including the tracker main process and attacker-to-victim transmission. The UAV identification module based on YOLO runs asynchronously, with an average latency of 288.6\,ms, but it does not block the main tracking or attack pipeline and therefore does not directly contribute to the redirection delay. The dominant cost in the online path comes from the tracker main process (235.5\,ms on average), while transmission remains lightweight (30.4\,ms). Despite this overhead, our experimental results show that the latency is sufficient for effective redirection in practice, and the framework remains robust even under temporary tracker losses, which will be discussed in Sec.~\ref{subsec:redirection_acc}.


\subsection{Redirection in Controlled Environments}
To verify the effectiveness of the spoofing capability enabled by the tracking pipeline, we first conducted indoor experiments in a closed arena using VICON motion capture system. The key metrics in this evaluation are: (1) redirection accuracy; (2) redirection covertness.

\subsubsection{Targeted Redirection Case Study}

\begin{figure*}[!t]
    \centering
\includegraphics[width=0.95\textwidth]{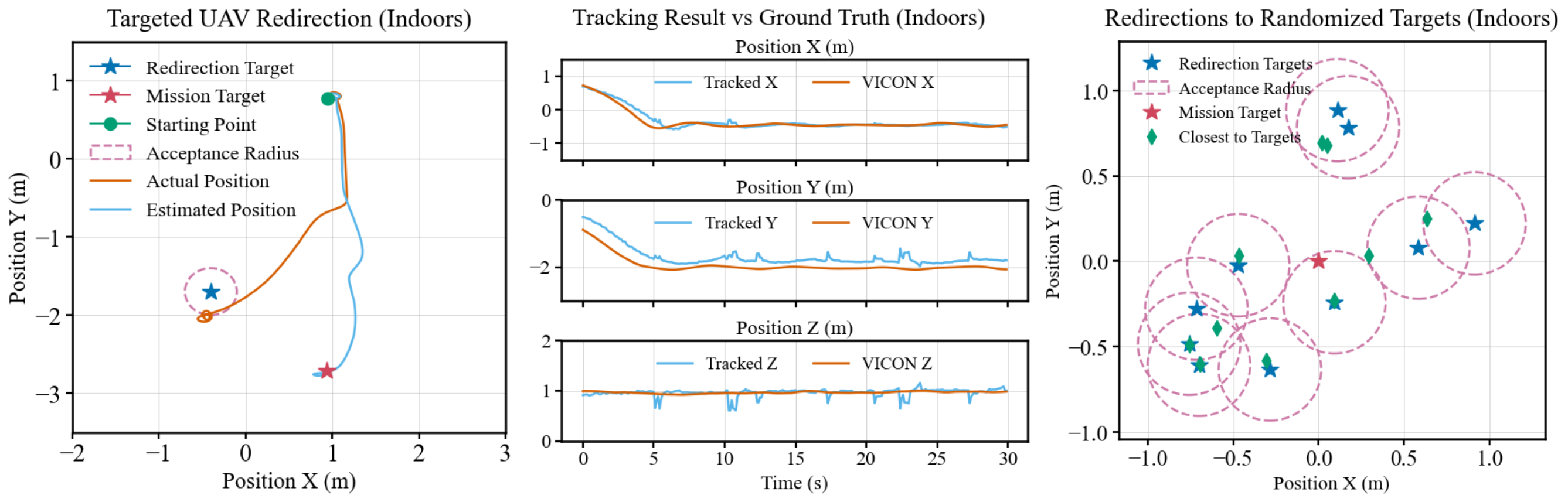}
    \caption{Real-time UAV redirection results indoors. Left: the actual and estimated position comparison of the victim drone in a targeted redirection; Middle: tracker performance in 3D positions; Right: redirection visualizations on 10 randomized targets.}
    \label{fig:vicon_redirection_results}
\end{figure*}

We first emulated a scenario where a victim drone is traveling along a certain trajectory toward a destination. The setup is designed such that: (1) The attacker tracks the real-time position of the victim; (2) The attacker computes the spoofing signal needed to redirect the victim to a designated area of attack, within a given acceptance radius; (3) The victim drone, with its GPS (emulated from VICON) spoofed, deviates from its original trajectory. As is shown in the left figure of Fig.~\ref{fig:vicon_redirection_results}, the victim is rerouted toward the redirection target while its onboard sensor fusion estimating itself still moving towards its original destination. The redirection algorithms managed to hold the victim drone at target and made it to land under the deception that its own destination had been achieved. To ensure covertness and strengthen the belief the victim is still moving towards the target, the redirection controller kept the victim drone in motion when: (1) The actual position has entered redirection area; (2) The estimated (deceived) position of the victim itself is still en-route to its original target. The final landing place is within the $0.3~m$ pre-defined acceptance radius for redirection. 

\subsubsection{Redirection Accuracy}
\label{subsec:redirection_acc}

\begin{table}[!t]
\centering
\renewcommand{\arraystretch}{1.1}
\caption{Redirection accuracy for indoor and outdoor attacks.}
\label{tab:redirection_accuracy_merged}

\begin{subtable}[t]{\columnwidth}
\centering
\caption{Indoor Redirection Position Error (m).}
\begin{tabularx}{\columnwidth}{*{5}{>{\centering\arraybackslash}X}}
\hline\hline
\textbf{Case 1} & \textbf{Case 2} & \textbf{Case 3} & \textbf{Case 4} & \textbf{Case 5} \\
\hline
0.057 & 0.007 & 0.003 & 0.165 & 0.295 \\
\hline
\textbf{Case 6} & \textbf{Case 7} & \textbf{Case 8} & \textbf{Case 9} & \textbf{Case 10} \\
\hline
0.161 & 0.282 & 0.008 & 0.211 & 0.057 \\
\hline\hline
\end{tabularx}
\label{tab:redirection_accuracy_cases}
\end{subtable}

\begin{subtable}[t]{\columnwidth}
\centering
\caption{Outdoor Redirection Position Error (m).}
\begin{tabularx}{\columnwidth}{c*{4}{>{\centering\arraybackslash}X}}
\hline\hline
\textbf{Case} &
\textbf{Hover} &
\textbf{Move I} &
\textbf{Move II} &
\textbf{Move III} \\
\hline
\textbf{1} & 0.658 & 0.044 & 0.351 & 0.382 \\
\hline 
\textbf{2} & 1.241 & 0.483 & 0.091 & 0.554 \\
\hline\hline
\end{tabularx}
\label{tab:outdoor_redirection_accuracy}
\end{subtable}

\end{table}

To further showcase the effectiveness of the redirection, we randomly sampled $10$ redirection destinations and conducted rerouting runs toward them. The right figure of Fig.~\ref{fig:vicon_redirection_results} demonstrates the targets (with their acceptance ranges), and the victim positions hitting the target. The redirection accuracy is shown in Table~\ref{tab:redirection_accuracy_cases}, and the mean redirection error is $0.125~m$ with the standard deviation at $0.107~m$. The fluctuation in the redirection accuracy is due to the tracking precision between varying tracker modes, where the availability of sensing sources is different.

\vspace{2pt}
\noindent\textbf{Redirection under Intermittent Tracker Loss.} While our tracker pipeline considers four different modes in response to different availability of the sensing sources, as discussed in Sec.~\ref{subsec:visual_lidar_pipeline}, intermittent tracker losses could still happen because of false positive detections, obstacle occlusion, or temporary loss of connection. However, despite the loss in tracker, the redirection controller still reserves the capability to produce estimated spoofing signals to ensure the continuity of redirection. More details are in App.~\ref{app:redirection_under_loss}.

\subsubsection{Attack Stealthiness Analysis}

\begin{figure*}[!t]
    \centering
\includegraphics[width=0.97\textwidth]{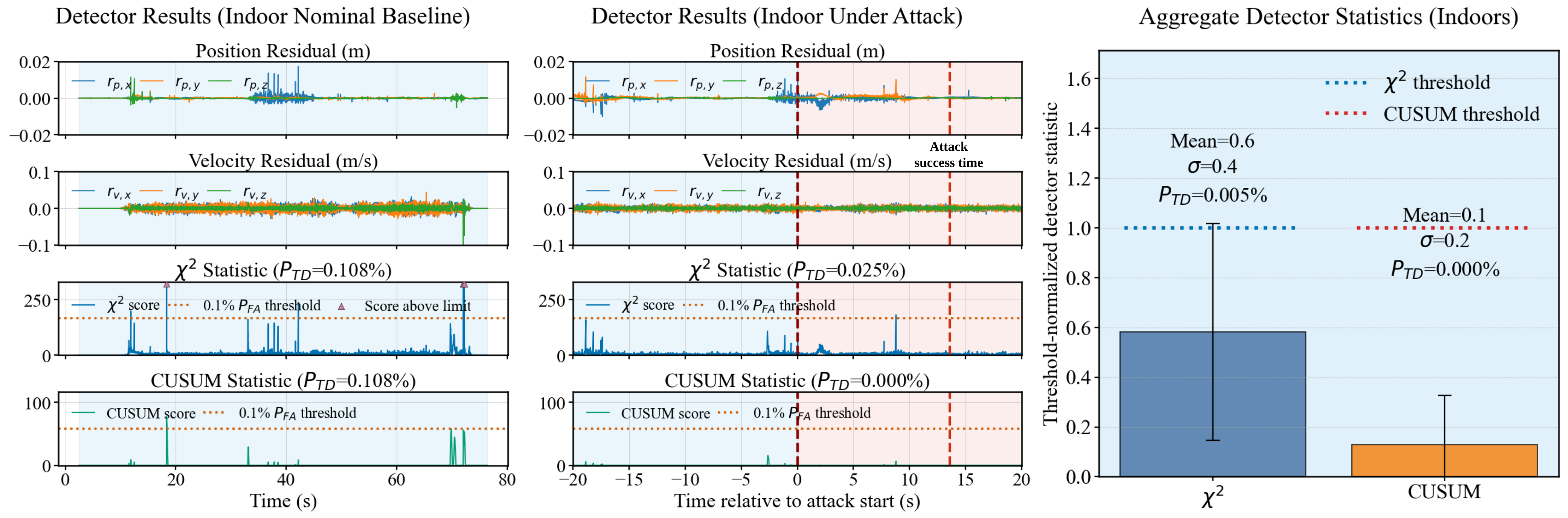}
    \caption{
    Anomaly detection results for indoor redirection attacks.
    Left: nominal flight with tuned detector thresholds.
    Middle: attack-aligned time-series response in the attack window with raw detector statistics in the case study. Right: aggregate detector statistics (threshold-normalized for comparing $\chi^2$ and CUSUM) over ten hovering attack trials.
    }
    \label{fig:indoor_detection}
\end{figure*}

As a key contribution of our paper, we focus on the anomaly detection on the real-world data. 

\vspace{2pt}
\noindent\textbf{Tuning Anomaly Detectors.} First, we implement the two benchmark anomaly detectors used in simulation, namely $\chi^2$ and CUSUM. As in the simulation setup, we \textit{tune their parameters to maintain a $0.1\%$ false positive rate} under nominal, non-attack operation.

\vspace{2pt}
\noindent\textbf{Real-world Stealthiness.}  Fig.~\ref{fig:indoor_detection} presents the detector responses in indoor experiments. In nominal flights, both detectors are operating normally at the designed false positive rate. Under attack, the case study exhibits negligible deviation during the redirection window, and neither detector shows a consistent increase strong enough to flag the attack ($0.025\%$ and $0.000\%$). Across ten attack trials, the threshold-normalized detector statistics remain well below the alarm boundary, with $\chi^2$ reaching a mean normalized score of $0.6$ and CUSUM only $0.1$. Correspondingly, the observed detection rates remain low, at $0.005\%$ for $\chi^2$ and zero detection for CUSUM. These results show that the proposed redirection attack \textit{remains highly stealthy against anomaly detectors} even on the physical UAV.

\subsection{Outdoor Redirection Verifications}

We also tested the redirection framework with the same devices outdoors, using data from real GPS receivers. Fig.~\ref{fig:exp_setup} depicts the outdoor experimental setup for the tracking and spoofing analysis, with attacker and victim both having Holybro M-10 GPS installed~\cite{holybro-m10-gps}. 

\begin{figure}[!t]
    \centering
    \includegraphics[width=\columnwidth]{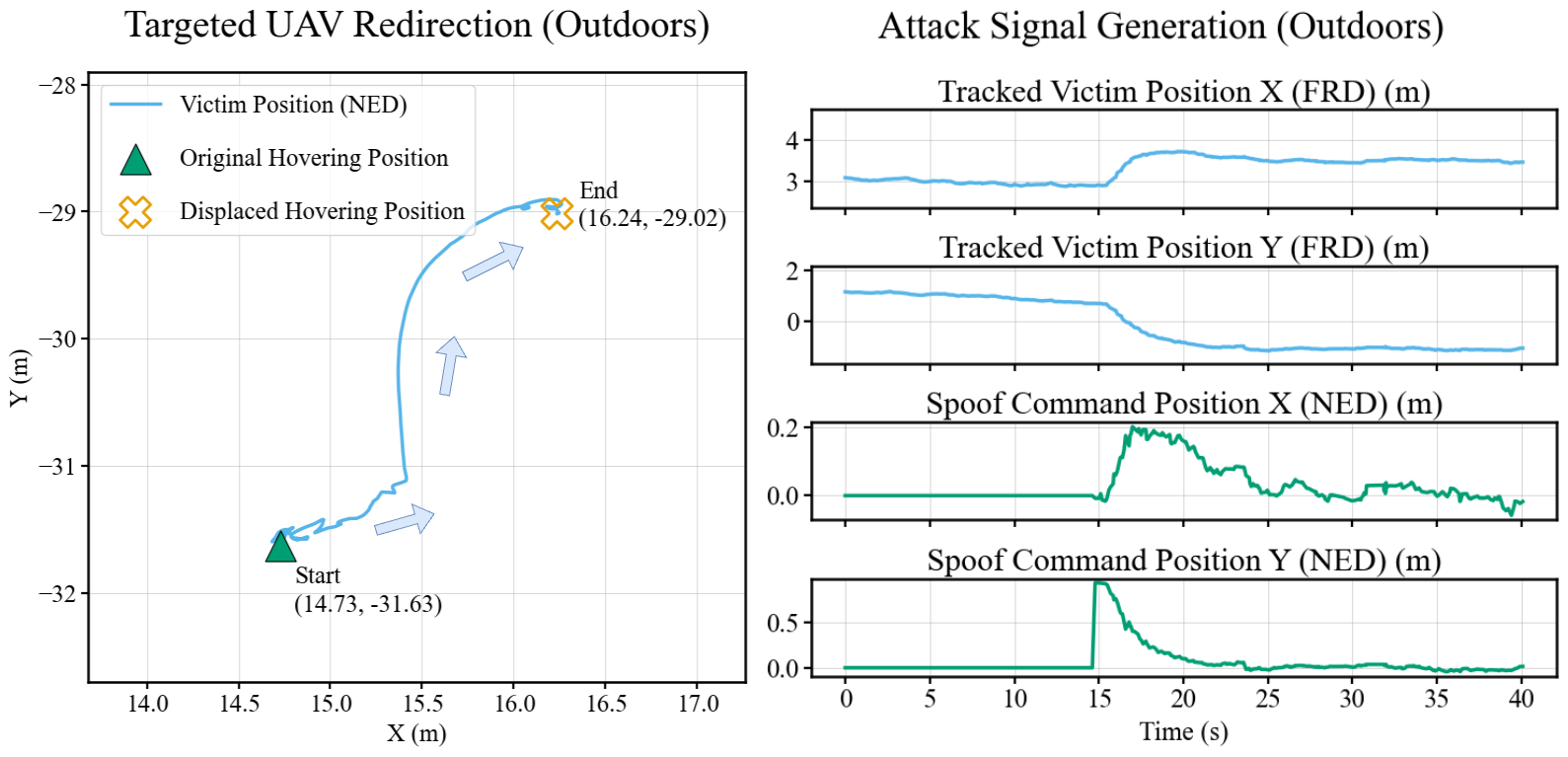}
    \caption{Outdoor redirection case study. Left: spoofed victim trajectory. Right: attacker side signal generation.}
    \label{fig:outdoor_hover_attack}
\end{figure}

\subsubsection{Outdoor Redirection Case Study}
We first conduct an outdoor hovering displacement experiment to demonstrate real-time UAV redirection in a GPS-enabled setting. As shown in Fig.~\ref{fig:outdoor_hover_attack}, the attacker estimates the victim's motion using the LiDAR-camera pipeline, and then computes spoofing commands using the tracked victim motion. These commands are transmitted to the victim and injected into its navigation pipeline, gradually biasing its position estimate and causing the UAV to drift from its original position to a displaced point without explicit awareness of the manipulation.

\subsubsection{Multi-Purpose Targeted Redirection} 

\begin{figure*}[!t]
    \centering
\includegraphics[width=0.98\textwidth]{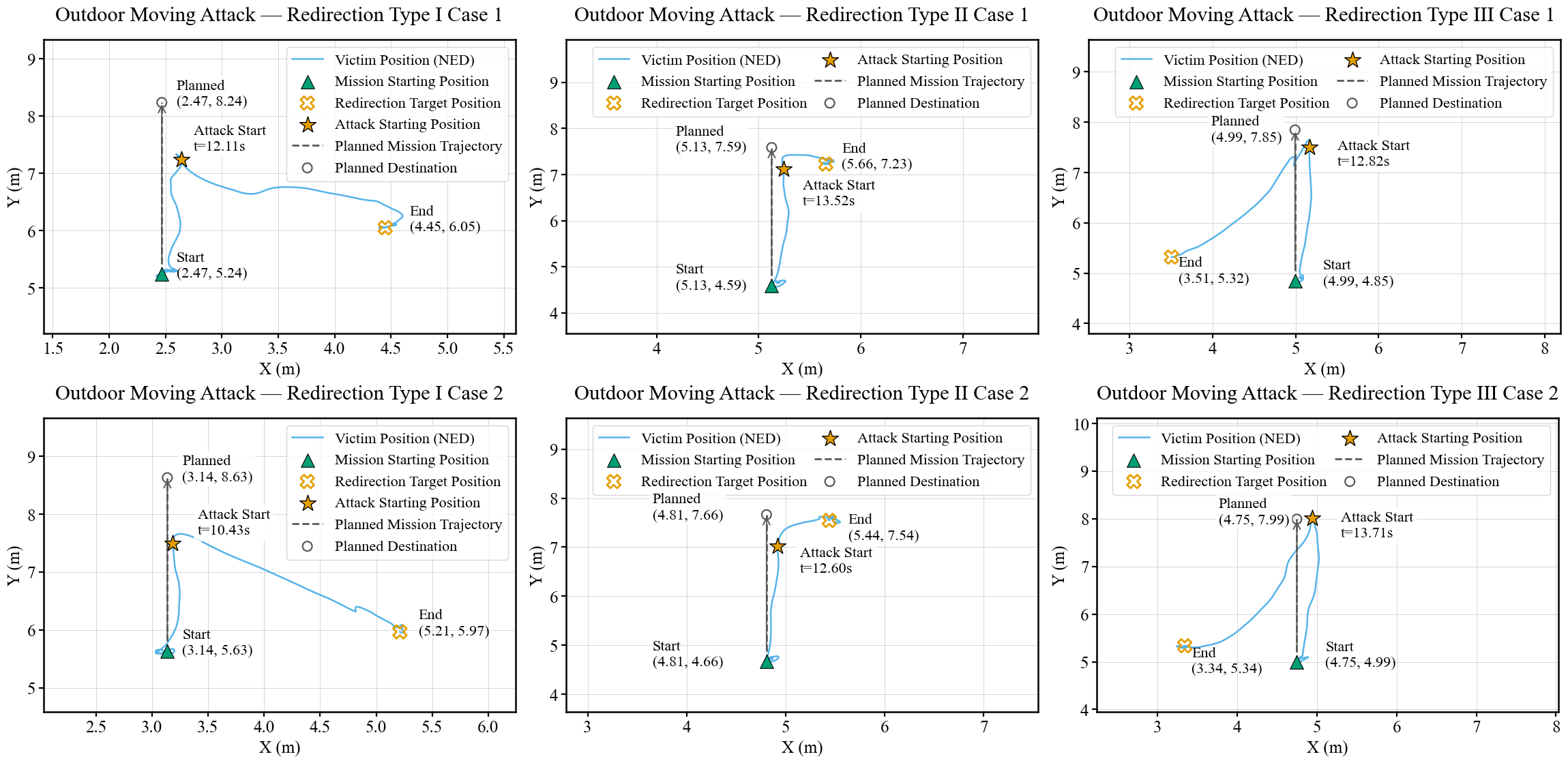}
    \caption{Outdoor targeted redirection against a moving victim UAV in the NED frame. Left: Type I---lateral redirection. Middle: Type II---halting redirection. Right: Type III---turn-back redirection. Each type is shown with two trials.}
    \label{fig:outdoor_move_attack}
\end{figure*}

We next evaluate outdoor targeted redirection in moving-flight missions. Using the same online redirection framework, the attacker can induce different victim behaviors simply by selecting different spoofing targets. Specifically, we consider three representative goals: \textit{Type I (Lateral Redirection)}, where the target is placed laterally to drive the victim away from its nominal path; \textit{Type II (Halting Redirection)}, where the target is chosen to arrest forward progress and hold the victim near the attack location; and \textit{Type III (Turn-back Redirection)}, where the target is placed behind the victim trajectory to induce a reversal of motion. Fig.~\ref{fig:outdoor_move_attack} shows these three qualitatively different outcomes are all achieved under the same attack pipeline.

The trajectories also show occasional irregularities caused by tracker mode switching, temporary tracking losses, and brief transmission drops. Nevertheless, the attack remains effective: once perception recovers, the framework resumes online target guidance and continues steering the victim toward the intended spoofing objective. Moreover, after reaching the redirected target, the victim proceeds on its own mission logic based on the corrupted navigation state, such as holding position or initiating mission completion behavior. Overall, these results show that the attacker can flexibly realize different goals by properly choosing spoofing targets, while remaining robust to perception and communication imperfections.

\subsubsection{Outdoor Redirection Accuracy and Stealthiness}

Table~\ref{tab:outdoor_redirection_accuracy} reports the redirection error as the distance between the victim position estimated by the LiDAR--camera tracking pipeline and the designated spoofing target. The mean redirection error is $0.476$\,m with a standard deviation of $0.351$\,m, which are approximately $3.8\times$ and $3.2\times$ those of the indoor experiments, respectively. Nevertheless, this error remains small relative to the noise level of standard GPS measurements, indicating that the proposed attack still achieves high practical precision in outdoor GPS-enabled flight.

\begin{figure*}[!t]
    \centering
\includegraphics[width=0.99\textwidth]{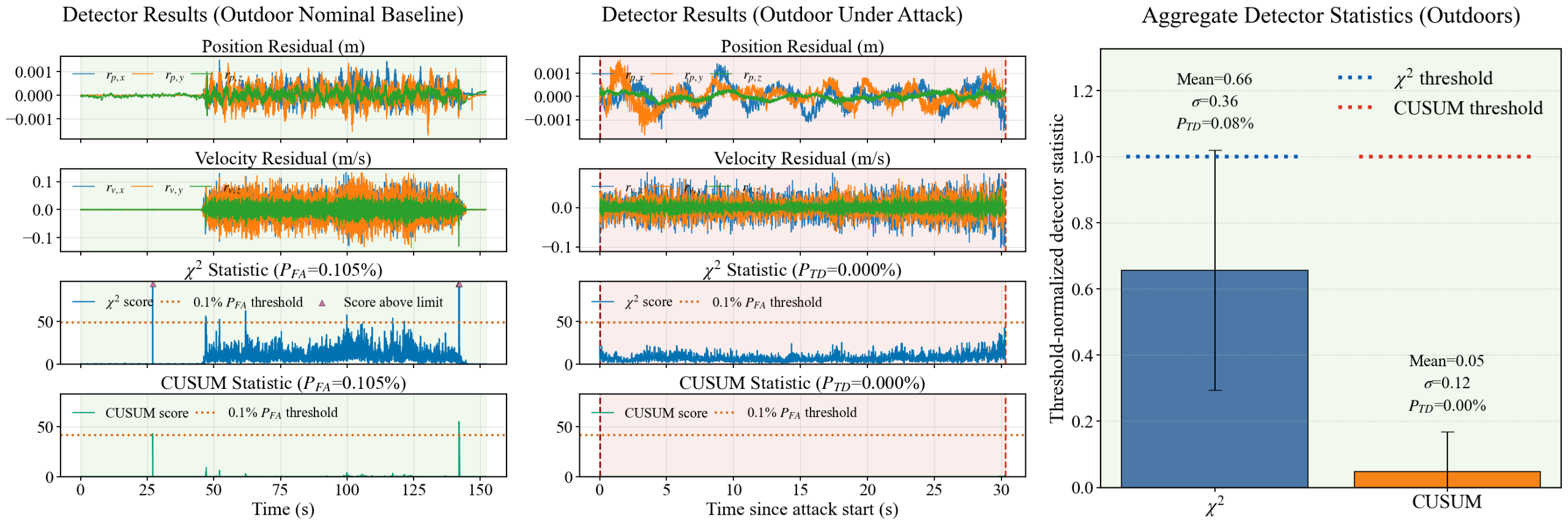}
    \caption{
    Anomaly detection results outdoors.
    Left: nominal response; Middle: attack case study; Right: aggregate statistics.
    }
    \label{fig:outdoor_detection}
\end{figure*}

\vspace{2pt}
\noindent\textbf{Stealthiness of Outdoor Attacks.} We deploy the same $\chi^2$ and CUSUM detectors, tuned under nominal outdoor flight for a $0.1\%$ false positive rate. Fig.~\ref{fig:outdoor_detection} shows both detectors are efficient under nominal conditions, with observed false positive rates of $0.105\%$. For the attack case study, the detector statistics do not reliably cross the alarm thresholds, leading to zero detections. Across all outdoor attacks, the aggregate detection rates are only $0.08\%$ and $0.00\%$, for $\chi^2$ and CUSUM respectively. These results show that the attack remains both precise and stealthy outdoors despite the increased sensing and environmental variability using real-world GPS.

\section{Discussion and Conclusion}
\label{sec:discussion}

This paper presents Phantom Navigator, a stealthy UAV redirection attack that exploits attacker-side tracking and GPS spoofing capabilities to redirect a victim UAV’s trajectory. The achievable redirection capacity is characterized by the strength of GPS spoofing, which can be efficiently estimated through pre-attack analysis. A real-time, lightweight redirection controller then generates spoofing signals based on the designated target and the estimated position of the victim drone, maintaining robustness against intermittent tracking degradation. Experimental evaluations on physical platforms validate the effectiveness of Phantom Navigator in achieving precise and reliable redirection in both indoor and outdoor environments. Next, we discuss the limitations, potential improvements, and defenses for our proposed UAV redirection attack framework:

\vspace{2pt}
\noindent\textbf{Stealthiness of Redirection Attacks.} 
\label{subsec:current_uav_defense}
In our work, the stealthiness of redirection is ensured by designing the attack signals satisfying the stealthiness constraints discussed in Sec.~\ref{subsec:redirection_model}. However, our stealthiness considers the worst-case scenario for the attacker: the victim is equipped with a highly efficient intrusion detection system that balances high detection accuracy with low time-to-detection, and is robust to mission and environment induced variations. Such detectors are not seen in commercial autopilots. Instead, they rely on static fail-safe strategies with overly conservative thresholds, leaving significant opportunities for stealthy spoofing. Therefore, the attacker can leverage the proposed stealthy redirection attack to induce substantial trajectory deviations while remaining below the detection threshold and satisfying the intended attack objectives.




\vspace{2pt}
\noindent\textbf{Toward a Moving Aerial Attacker Platform.}
In this work, we consider a stationary attacker platform. Extending the attack to a moving aerial platform is an important future direction, but it introduces additional challenges for the attacker. A key issue is \emph{self-spoofing}: when transmitting GPS spoofing signals, the attacker may corrupt its own GPS measurements. To mitigate this effect, a moving attacker could rely on independent localization sources, including $(i)$ multi-GNSS fusion across constellations~\cite{csahin2017optimal, ni2022gnss}, $(ii)$ onboard LiDAR- and camera-based SLAM~\cite{zhu2023camera}, and $(iii)$ other signals of opportunity (SOP)~\cite{9732682}. In contrast, under a \emph{stealthy} attack, the victim cannot reliably leverage such cross-checks, as the induced deviations remain within nominal noise bounds, making it difficult to identify which localization source has been compromised.

\vspace{2pt}
\noindent\textbf{Improving Tracking and Redirection Range.}
While this paper presents a LiDAR-camera tracking pipeline that is computationally lightweight, and robust under various levels of sensor degradation, its performance remains inherently constrained by the sensing range and resolution of the camera and LiDAR. In scenarios of prolonged sensory loss, the filter operating in prediction-only mode may gradually drift due to the absence of measurement updates. A potential extension is to integrate RADAR into the pipeline, offering long-range perception that complements the mid-range coverage of LiDAR and the high-resolution but short-range nature of cameras.

\vspace{2pt}
\noindent\textbf{Defenses against Malicious Redirection.}
Potential defenses include two directions. First, higher-accuracy positioning systems, such as RTK-GPS, can reduce the localization uncertainty and limit the magnitude of stealthy spoofing signals. Second, incorporating environmental semantics or leveraging collaborative sensing from trusted agents and infrastructure can establish trust across different onboard perception modules and help identify compromised localization sources under malicious redirection attacks.


\begin{acks}
This work is sponsored in part by the ONR under agreement N00014-23-1-2206, AFOSR under the award number FA9550-19-1-0169, and by the NSF under NAIAD Award 2332744 as well as the National AI Institute for Edge Computing Leveraging Next Generation Wireless Networks, Grant CNS-2112562. Moreover, the research was sponsored by the Army Research Office and was accomplished under Cooperative Agreement Number W911NF-26-2-A165. The views and conclusions contained in this document are those of the authors and should not be interpreted as representing the official policies, either expressed or implied, of the Army Research Office or the U.S. Government. The U.S. Government is authorized to reproduce and distribute reprints for Government purposes notwithstanding any copyright notation herein.
\end{acks}

\bibliographystyle{ACM-Reference-Format}
\bibliography{reference}

\appendix
\section{Appendix}

\subsection{Modeling Vehicle Physical Dynamics}
\label{app:uav_physical_dynamics}
To better illustrate the effect of GPS spoofing to achieve redirection, we discuss the vehicle dynamical model used in our formulations. The UAV physical dynamics can be captured with a standard nonlinear quadcopter model~\cite{sun2022comparative}. The translational motion 
is described by a standard Euler-Newton equation~\cite{sun2022comparative}:
\begin{equation}
\label{eq:model1}
\ddot{\mathcal{\xi}} = (\mathbf{R}\mathbf{f}^{\mathcal{B}} + \mathbf{f}_a)/m+\mathbf{g},
\end{equation}
where $\mathcal{\xi}=\begin{bmatrix}x,y,z\end{bmatrix}^T$ captures the coordinates of the UAV, $m$ denotes 
the total mass of the vehicle, and $\mathbf{g}=\begin{bmatrix}0,0,g\end{bmatrix}^T$ is the gravity acceleration vector; $\mathbf{f}^{\mathcal{B}}=\begin{bmatrix}0,0,T\end{bmatrix}^T$ is the force vector in the body frame for the collective thrust $T$ applied at the center of mass; $\mathbf{f}_a$ is the aerodynamic drag force in the Earth frame
%
$\mathbf{f}_a = 
\left[\begin{smallmatrix}
-k_{d,x} v_x,
-k_{d,y} v_y,
-k_{d,z} v_z + k_h \left(v_x^2 + v_y^2 \right)
\end{smallmatrix}\right]^T;$ here, $k_{d,x},k_{d,y}, k_{d,z},k_h$ are drag force coefficients and $v_x, v_y, v_z$ are the earth-frame velocities. 

The rotational motion can be captured as
\begin{equation}
\label{eq:model2}
\dot{\mathbf{R}}=\mathbf{R}\hat{\bm{\Omega}}^{\mathcal{B}}, \quad
\bm{\mathcal{I}}\dot{\bm{\Omega}}^{\mathcal{B}}=-\bm{\Omega}^{\mathcal{B}}\times \bm{\mathcal{I}}{\bm{\Omega}}^{\mathcal{B}} + \bm{\tau}^{\mathcal{B}},
\end{equation}
where $\bm{\Omega}^{\mathcal{B}}=\begin{bmatrix}\Omega_x,\Omega_y,\Omega_z\end{bmatrix}^T$ is the angular velocity in the body frame, $\hat{.}$ denotes the operator that maps a vector in $\mathbb{R}^3$ to a skew-symmetric matrix, $\bm{\mathcal{I}}\in \mathbb{R}^{3\times 3}$ is the inertia matrix, and
$\bm{\tau}^{\mathcal{B}}=\begin{bmatrix}\tau_x,\tau_y,\tau_z\end{bmatrix}^T$ 
is the total torque vector in the body~frame.

\subsection{GPS Spoofing Effect Formulation}
\label{app:gps_spoofing_math}
In a standard UAV sensor fusion configuration, for which the model of the physical dynamics is provided in App.~\ref{app:uav_physical_dynamics}, GPS provides direct measurements of the earth-frame position and velocity:
\begin{equation}
\mathbf{y}^{\text{GPS}}_k = 
\begin{bmatrix}
\mathbf{p}_k \\
\mathbf{v}_k
\end{bmatrix}
+
\begin{bmatrix}
\mathbf{n}^{\text{pos}}_k \\
\mathbf{n}^{\text{vel}}_k
\end{bmatrix}, \quad
\mathbf{n}^{\text{pos}}_k \sim \mathcal{N}(0, \Sigma_p),\ 
\mathbf{n}^{\text{vel}}_k \sim \mathcal{N}(0, \Sigma_v);
\label{eq:gps_obs_full}
\end{equation}
here $\mathbf{y}^{\text{GPS}}_k$ captures GPS sensor readings, modeled as the true position and velocity $\mathbf{p}_k,\mathbf{v}_k$, and the zero-mean Gaussian noise $\mathbf{n}^{\text{pos}}_k, \mathbf{n}^{\text{vel}}_k$. As discussed in GPS takeover attacks, GPS receivers lock onto counterfeit GPS signals with offsets added to the original sensor readings:
\begin{equation}
\mathbf{y}^{\text{GPS,s}}_k =
\begin{bmatrix}
\mathbf{p}_k \\
\mathbf{v}_k
\end{bmatrix}
+
\begin{bmatrix}
\mathbf{n}^{\text{pos}}_k \\
\mathbf{n}^{\text{vel}}_k
\end{bmatrix}
+
\begin{bmatrix}
\mathbf{s}^{\text{pos}}_k \\
\mathbf{s}^{\text{vel}}_k
\end{bmatrix},
\label{eq:gps_obs_spoofed_full}
\end{equation}
where $\mathbf{y}^{\text{GPS,s}}_k$ denotes the spoofed GPS readings received by the UAV controller; and $\mathbf{s}^{\text{pos}}_k, \mathbf{s}^{\text{vel}}_k$ correspond to the spoofing signals added to position and velocity, respectively. These spoofed terms effectively bias the GPS output, thus leading to incorrect state estimation in the sensor fusion process:
\begin{equation}
\hat{\mathbf{x}}^s_k = g(\hat{\mathbf{x}}_{k-1}, \mathbf{u}_{k-1}, \mathbf{y}^s_k)
\label{eq:fusion}
\end{equation}
where $\hat{\mathbf{x}}^s_k$ stands for the corrupted state estimate from sensor fusion process $g$ due to spoofed sensor measurements $\mathbf{y}^s_k=\begin{bmatrix}
\mathbf{y}^{\text{GPS,s}}_k,\mathbf{y}^{\text{IMU}}_k
\end{bmatrix}$. 
Note that the IMU measurements are assumed to be uncompromised in this work. 
Previously, it was shown that attacking GPS measurements only is sufficient to spoof the position of the UAV in a stealthy way, even without accessing the IMU readings of the drone~\cite{10885819,khazraei_cdc22}.

\subsection{Simulation Setup}
\label{app:simulation_setup}

The simulation parameters of MATLAB/Simulink are provided in Table~\ref{tab:sim-params}, including the simulator configurations, redirection commands setup, drone physical model, and noise profiles.

\begin{table}[!t]
\centering
\small
\renewcommand{\arraystretch}{1.15}
\caption{Simulation and drone parameters used in MATLAB/Simulink for offline verification.}
\label{tab:sim-params}
\begin{adjustbox}{width=\columnwidth}
\begin{tabular}{p{0.55\columnwidth} p{0.42\columnwidth}}
\toprule
\textbf{Parameter Groups} & \textbf{Value} \\
\midrule

\multicolumn{2}{l}{\textbf{Simulator}} \\
\hline
Simulation duration & 400 s \\
Sample time & 0.01 s \\
\midrule

\multicolumn{2}{l}{\textbf{Redirection Command Station}} \\
\hline
Position tracking noise & 0.5 m (std. dev.) \\
Velocity tracking noise & 0.1 m/s (std. dev.) \\
Redirection strength (acceleration) & 1.0 m/s\textsuperscript{2} \\
\midrule

\multicolumn{2}{l}{\textbf{Drone Physical Model}} \\
\hline
Moments of inertia ($I_{xx},I_{yy},I_{zz}$) & 0.016, 0.016, 0.0274 kg·m\textsuperscript{2} \\
Rotor inertia & $3.789\times10^{-6}$ kg·m\textsuperscript{2} \\
Mass & 1 kg \\
Gravity & 9.8 m/s\textsuperscript{2} \\
\midrule

\multicolumn{2}{l}{\textbf{Sensor and System Noise}} \\
\hline
GPS position noise & 0.5 m (std. dev.) \\
GPS velocity noise & 0.01 m/s (std. dev.) \\
IMU accelerometer noise & 0.035 m/s\textsuperscript{2} \\
IMU gyro noise & 0.021 rad/s \\
IMU magnetometer noise & 0.206 (normalized)\\
\bottomrule
\end{tabular}
\end{adjustbox}
\end{table}

\subsection{Redirection under Tracker Loss}
\label{app:redirection_under_loss}

\begin{figure}[!t]
    \centering
\includegraphics[width=0.98\columnwidth]{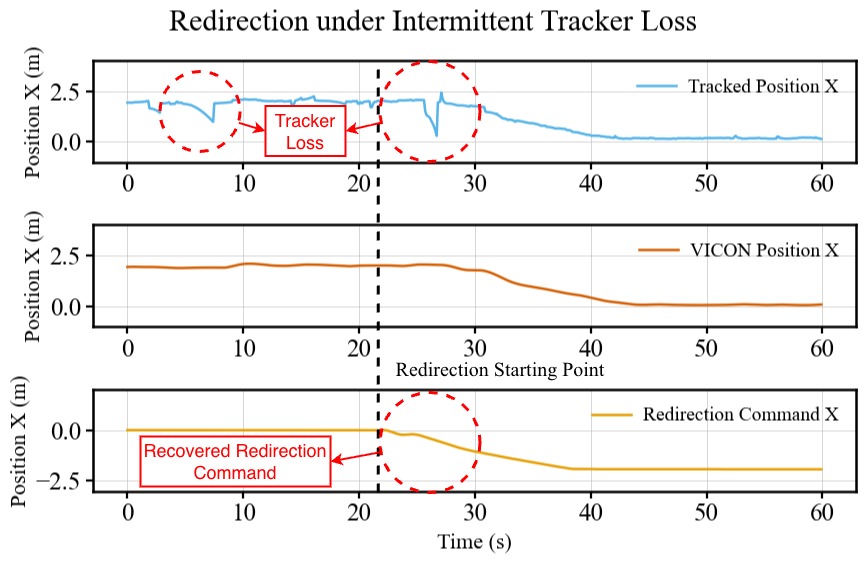}
    \caption{Redirection result in X-axis under intermittent tracker loss with recovered redirection command.}
    \label{fig:vicon_redirection_tracker_loss}
\end{figure}

We evaluated the performance of our LiDAR-camera tracker when intermittent target loss takes place. As is shown in Fig.~\ref{fig:vicon_redirection_tracker_loss}, tracker loss happens both before and after the beginning of redirection operation, with very large tracking errors in place. Nevertheless, the redirection controller is not significantly impacted and diverts the victim drone coherently, proving its robustness to these errors.

\end{document}